\documentclass[pdflatex,sn-mathphys]{sn-jnl}

\theoremstyle{thmstyleone}%
\theoremstyle{thmstyletwo}%

\theoremstyle{thmstylethree}%

\begin{document}

\title[Disparity Refinement for Learning-Based Methods]{A Disparity Refinement Framework for Learning-based Stereo Matching Methods in Cross-domain Setting for Laparoscopic Images}

\author*[1]{\fnm{Zixin} \sur{Yang}}\email{yy8898@rit.edu}

\author[2]{\fnm{Richard} \sur{Simon}}\email{rasbme@rit.edu}

\author[1,2]{\fnm{Cristian} \sur{A. Linte}}\email{calbme@rit.edu }

\affil[1]{\orgdiv{Center for Imaging Science}, \orgname{Rochester Institute of Technology}, \orgaddress{\street{1 Lomb Memorial Dr}, \city{Rochester}, \postcode{14623}, \state{NY}, \country{USA}}}

\affil[2]{\orgdiv{Biomedical Engineering}, \orgname{Rochester Institute of Technology}, \orgaddress{\street{1 Lomb Memorial Dr}, \city{Rochester}, \postcode{14623}, \state{NY}, \country{USA}}}


\abstract{\textbf{Purpose:} Stereo matching methods that enable depth estimation are crucial for visualization enhancement applications in computer-assisted surgery (CAS). Learning-based stereo matching methods are promising to predict accurate results on laparoscopic images. However, they require a large amount of training data, and their performance may be degraded due to domain shifts.  

\textbf{Methods:} Maintaining robustness and improving the accuracy of learning-based methods are still open problems. To overcome the limitations of learning-based methods, we propose a disparity refinement framework consisting of a local disparity refinement method and a global disparity refinement method to improve the results of learning-based stereo matching methods in a cross-domain setting. Those learning-based stereo matching methods are pre-trained on a large public dataset of natural images and are tested on two datasets of laparoscopic images.

\textbf{Results:}  Qualitative and quantitative results suggest that our proposed disparity framework can effectively refine disparity maps when they are noise-corrupted on an unseen dataset, without compromising prediction accuracy when the network can generalize well on an unseen dataset.

\textbf{Conclusion:} Our proposed disparity refinement framework could work with learning-based methods to achieve robust and accurate disparity prediction. Yet, as a large laparoscopic dataset for training learning-based methods does not exist and the generalization ability of networks remains to be improved, the incorporation of the proposed disparity refinement framework into existing networks will contribute to improving their overall accuracy and robustness associated with depth estimation.
}


\maketitle

\section{Introduction}\label{sec1}



Stereo endoscopy is commonly used to enable depth estimation in laparoscopic surgery\cite{mountney2010three, lin2016video, allan2021stereo,edwards2020serv}. Stereo correspondences represented by disparity maps can be estimated via stereo matching techniques \cite{scharstein2002taxonomy} to provide depth measurements with known intrinsic and extrinsic camera calibration.

Depth information offered by stereo matching can play a key role in surgical navigation \cite{lin2016video} and visualization enhancement applications \cite{mountney2010three,modrzejewski2019vivo}.
Depth information can improve surgical performance by enabling the tracking of tissue surface deformations for rendering a motion-stabilized view and avoiding surgical instrument collisions with critical anatomical structures \cite{mountney2010three,lin2016video}. Depth estimation also benefits downstream tasks, such as 3D surface reconstruction/SLAM \cite{zhou2019real,recasens2021endo,wei2022stereo,yang2022endoscope}, and registration \cite{modrzejewski2019vivo} algorithms, which are the foundations of advanced applications in compute-integrated intervention \cite{mountney2010three,lin2016video,modrzejewski2019vivo}.

Over the last decades, an extensive range of stereo matching approaches has been presented, divided into traditional and deep learning-based methods. However, these methods still have their shortcomings when it comes to handling stereo endoscopic images with smooth surfaces, specular highlights, patterns of repetition, and uneven illumination.

Traditional methods consist of local \cite{scharstein2002taxonomy,bleyer2011patchmatch} and global \cite{scharstein2002taxonomy,terzopoulos1986regularization} methods, according to the classifications in \cite{scharstein2002taxonomy}. Local methods use a pre-defined search range to determine the optimal disparity map. In contrast, global methods use the entire image to formulate an optimization problem with a data term and a regularization term. Local methods are computationally efficient, but they have difficulty handling feature-less surfaces and specular highlights, which are common in endoscopic images. Global methods can provide relatively accurate results; however, they can be time-consuming and may require good initialization, especially for large displacements \cite{zach2007duality,horn1981determining}.

Learning-based stereo matching methods can be roughly divided into fully-supervised methods \cite{chang2018pyramid,xu2020aanet} and self-supervised methods \cite{li2021sins,yang2021dense}. The former uses accurate ground truth disparity maps for training, while the latter relies on formulating image synthesis loss. In the era of deep learning, learning-based stereo matching methods are reported to achieve high performance on several public benchmark datasets and outperform traditional methods \cite{flow,kitti}. However, their achievements are based on several prerequisites: 1. a large amount of data available for training; 2. identically distributed training and testing datasets. Those prerequisites are not generally satisfied in the computer-assisted surgery application field. Obtaining a sufficiently large dataset to train learning-based methods that can require accurate ground truth, usually with millions of parameters, is impractical in the surgical field. Given various texture and surgery settings, there is also no guarantee that the distribution difference between the training dataset and the testing dataset is negligible. As a large laparoscopic dataset with accurate ground truth does not exist, several methods \cite{li2021revisiting,long2021dssr,edwards2020serv} use cross-domain dataset \cite{kitti,flow} for training, which may be not robust. Hence, the distribution change or domain shift can jeopardize the performance \cite{wang2018deep}.   


Considering the advantages and disadvantages of the mentioned methods, a motivating question arises: Can we combine learning-based methods with traditional methods to achieve more robust and accurate results?




To answer the question, in this paper, we study the use of traditional methods to refine disparity maps from learning-based methods in a cross-domain setting. We propose a refinement framework for disparity maps of endoscopic images predicted from learning-based methods trained on a large stereo dataset of natural images. The proposed disparity refinement framework consists of local and global refinement methods (LDR and GDR).

We first use the local refinement method (LDR) to detect low confidence regions on the predicted disparity map by assuming that outliers in the disparity map strongly violate the smoothness and photometric consistency assumptions. Next, the disparity values of low confidence regions are interpolated from the surrounding high confidence regions. Subsequently, we introduce a multi-resolution variational model using a proposed illumination invariant data term in the global refinement method to further refine the disparity from the previous step. The model uses the refined disparity map from the first stage as initialization and is implemented on the GPU, which improves its disparity refinement accuracy, robustness, and computational efficiency. The proposed method is evaluated on two laparoscopic datasets. Experimental results demonstrate that 1) the proposed method can effectively improve the performance of several learning-based methods when they pose difficulties in generalizing on an unseen dataset. 2) it does not impair prediction performance when the learning-based methods can generalize well to an unseen dataset. 3) our framework has better performance in terms of accuracy and speed compared to the other recent, closely related method.

Our contributions are summarized as follows:

\begin{itemize}

\item We present a disparity refinement framework based on traditional methods for learning-based stereo matching methods consisting of LDR and GDR methods in a cross-domain setting. 

\item We present an LDR method to measure the confidence of disparity maps and refine disparity values in low confidence regions. The method is designed to refine errors concentrated in small regions and provide a more robust initialization for the subsequent global disparity refinement method. 

\item We present a GDR method using our illumination invariant multi-resolution variational model to refine various artifacts, especially for errors concentrated in large regions.

\end{itemize}







\section{Related Works}

Stereo matching is a classical problem in computer vision with a large number of methods that have been proposed. Readers may refer to benchmarks \cite{kitti,flow} for the list of stereo matching methods. We review the most relevant works to our applications.

Local matching methods are favored by applications that require real-time processing. Stoyanov et al. \cite{stoyanov2010real} presented a local stereo matching method that propagates disparity information around feature matches in salient regions to avoid mismatches in specular highlights and occlusions. ELAS \cite{geiger2010efficient}, one of the widely used stereo matching algorithms in surgical scenes, used feature matches to generate a prior pixel-wise disparity and then optimize the disparity via the maximum a-posteriori algorithm. 

Global methods using variational approaches are attracting attention, as they are robust to occlusions and specular highlights. Global methods can be accompanied by a sparse to dense step to initialize global methods. In the Dense inverse search (DIS) algorithm \cite{kroeger2016fast}, it first searches for patch correspondences and then creates a dense displacement field through patch aggregation, which is then fed to a variational model. Song {\it et al.} \cite{song2021bayesian} proposed a Bayesian framework to improve the patch aggregation step in DIS, which improves its performance on textureless surfaces and the photometric inconsistency. Xia {\it et al.} \cite{xia2022robust} proposed a method that includes a sparse-dense feature matching step, image illumination equalization, and global variational refinement. Using deep learning to provide acceptable initialization is also explored in \cite{revaud2015epicflow,roxas2018real}.


Different neural network architectures have also been exploited to achieve accurate disparity prediction. PSMnet \cite{chang2018pyramid} uses 3D convolutions to the aggregate global context information and is the winning technique in the grand challenge of endoscopic data \cite{allan2021stereo}. AAnet \cite{xu2020aanet} replaced 3D convolutions with cross-scale correlation, achieving faster and better performance. LEAStereo \cite{cheng2020hierarchical} utilized a framework to optimize the architecture of the stereo matching network automatically. The LEAStereo achieved top accuracy in several benchmarks \cite{kitti,flow}.

Work related to improving the generalization ability of learning-based stereo matching methods is relatively less exploited. Instead of using learning-based features, Cai {\it et al.} \cite{cai2020matching} uses hand-crafted features in the matching space to keep the network from learning domain-specific features. Li {\it et al.} \cite{li2021revisiting} introduced a stereo matching neural network with the Transformer architecture \cite{vaswani2017attention}, and demonstrated the network could generalize across different domains. Pipelines that adapt networks to unseen target domains are also proposed \cite{song2021adastereo,bousmalis2017unsupervised,mahmood2018unsupervised}. However, those pipelines require samples from the target domain, limiting their applications to small datasets, such as medical images.


Disparity refinement is usually a final step in traditional stereo matching methods \cite{scharstein2002taxonomy}. Disparity maps can be directly refined by removing peaks \cite{ma2013constant}, smoothing filters \cite{ma2013constant}, and hole filling \cite{hirschmuller2003stereo, zhou2019real}. Most refinement methods detect outliers and then correct them to achieve more accurate results. In \cite{hirschmuller2005accurate}, outliers are detected via left-right consistency (LRC) check and refined with steps including peak removal, interpolation, and median filtering. Banno and Ikeuchi \cite{banno2011disparity} refined outliers failed LRC check with the proposed directed anisotropic diffusion technique. LRC is one of the most common methods to detect outliers. However, LRC doubles the computational time of traditional methods and does not apply to learning-based methods, as most existing learning-based methods are trained to predict only the disparity map of the left image. Zhan {\it et al.} \cite{zhan2015accurate} proposed a multistep refinement method to classify outliers and recover the outliers according to their kind. Assuming the tissue surface is relatively smooth, Zhou and Jagadeesan \cite{zhou2019real} introduced a radius-based outlier removal method, followed by holing and smoothing. However, those methods can not refine large areas with errors.


A few works have proposed refinement methods for learning-based methods. To our best knowledge, the closest work is \cite{yan2019segment}, which refines the disparity map from a single method. 

Our work focuses on a disparity refinement method robust to different learning-based methods in a cross-domain setting.

\section{Methods}\label{sec11}


\noindent We focus on leveraging reasonable assumptions and traditional methods to correct errors in disparity maps predicted by several state-of-the-art learning-based stereo matching methods in a cross-domain setting. We use their publicly released models, trained on a large public dataset, to make predictions on laparoscopic image datasets. Disparity maps from the learning-based methods may contain quite a few errors, resulting from the difference between the training dataset and the testing dataset and the properties of endoscopic images. We propose a disparity refinement framework consisting of a local disparity refinement (LDR) method and a global disparity refinement (GDR). 
\subsection{Local Disparity Refinement}

We use a detection and correction strategy in the local disparity refinement stage. We first estimate the disparity map's confidence map, and a threshold is set to select outliers from the confidence map. Finally, the disparity values of outliers are interpolated from neighboring pixels.




Several assumptions are made to estimate the confidence of the disparity. Firstly, we assume that the tissue surface is relatively smooth. The pixel $\mathbf{x}$ should have high confidence in a smoothness confidence map $C_{s}(\mathbf{x})$ if the disparity value $u(\mathbf{x})$ is consistent with its surrounding pixels $\overline{u}_{w}(\mathbf{x})$:

\begin{align} \label{eq::1}
    C_{s}(\mathbf{x}) =  1 - \alpha_s \cdot \bigg\lvert\frac{u(\mathbf{x})-\overline{u}_{w}(\mathbf{x})}{\overline{u}_{w}(\mathbf{x})}\bigg\lvert,
\end{align}

\noindent where $\overline{u}_{w}(\mathbf{x})$ is the mean disparity value of the local window with the size $w$, and 
 $\alpha_s$ is the hyper-parameter.

Secondly, we assume that outliers in the disparity map tend to violate the photo-consistency assumption strongly. Intensities of outliers $I_s(\mathbf{x})$ in the source (left) image and their matched points $I_{t}(\mathbf{x}+u(\mathbf{x}))$ in the target (right) image would have a large difference. Due to illumination differences, the intensity values of corresponding images may not be the same. However, it is reasonable that outliers would strongly violate this assumption. Therefore, the photo-consistency confidence $C_{p}(\mathbf{x})$ is defined as:

\begin{align} \label{eq::2}
C_{p}(\mathbf{x}) =  1 -\alpha_p \cdot \bigg\lvert\frac{I_{s}(\mathbf{x})-{I}_{t}(\mathbf{x}+u(\mathbf{x}))}{I_{s}(\mathbf{x})}\bigg\rvert,
\end{align}

\noindent where $\alpha_p$ is the hyper-parameter. Pixels with incorrect disparity values have low confidence values in  $C_{p}(\mathbf{x})$.


Thirdly, we assume that in specular highlights and border occlusions \cite{huq2013occlusion}, predicted disparities would tend to be unreliable. In specular highlights, pixel intensities are saturated and uniform. Border occlusions result from the fact that the right camera misses some of the leftmost portions of the field of view of the left camera.

We set confidence values in these regions as zeros by introducing the specular highlight mask $M_s(\mathbf{x})$ and the boundary occlusions mask $M_b(\mathbf{x})$: 

\begin{align}\label{eq::3}
   M_s(\mathbf{x}) =
\begin{cases}
1 & \text{if  }  S(x)>th_s \\
0 & \text{otherwise.} \\
\end{cases}
\end{align}

\begin{align}\label{eq::4}
   M_b(\mathbf{x}) =
\begin{cases}
1 & \text{if  }  \mathbf{x}+D(\mathbf{x}) \ \text{exists in the right image} \\
0 & \text{otherwise.} \\
\end{cases}
\end{align}

\noindent where $S(x)\in[0,1]$ is the value in HSV color space.

The final confidence map $C_f(\mathbf{x})$ is defined as the product of the above confidence maps and masks:

\begin{align} \label{eq::5}
C_f(\mathbf{x}) =  M_v(\mathbf{x}) \cdot M_s(\mathbf{x}) \cdot C_p(\mathbf{x})\cdot C_{s}(\mathbf{x}) .
\end{align}


Pixels are selected as outliers if their confidence values are below
the threshold value of $th_f$. Next, we search for inliers along with eight directions for each outlier, similar to SGM \cite{hirschmuller2003stereo}. Then, for each outlier, its disparity is replaced with the median value of the eight inliers.

\subsection{Global Disparity Refinement}


We formulate stereo matching as a variational problem. Disparity values ${u}$($x$) between the source $I_{s}$ and target $I_{t}$ images are predicted by the minimization of an energy function composed of a data term $E_{data}$, and a regularization term $E_S$:

\begin{align} \label{eq::6}
    \min_u \left[\lambda \,E_{data}({u},I_{t},I_{s})+ E_S({u}) \right ],
\end{align}
\noindent where $\lambda$ denotes the weight between $E_{data}$ and $E_S$. 
$E_{data}$ measures the similarity of pixels in $I_{s}$ and $I_{t}$ using:

\begin{align} \label{eq::7}
E_{data}({\bf u}) = \int_\Omega \left\lvert {\bf D}(P(x+u,I_t))-{\bf D}(P(x,I_s)) \right\lvert^2 dx.
\end{align}

\noindent Here, $\Omega$ denotes the image domain, $x$ presents the pixel location, $P(x,I)$ is the patch that contains the local intensities, and $\mathbf{D}$ is a novel illumination invariant descriptor:

\begin{align} \label{eq::8}
    P(x)=\begin{bmatrix}
I(x_4) & I(x_3) & I(x_2)\\
I(x_5) & I(x) & I(x_1)\\
I(x_6) & I(x_7) & I(x_8)\end{bmatrix}.
\end{align}

\begin{align} \label{eq::9}
{\bf D}(P(x_0,I)) = \frac{\mathbf{A}(P(x_0,I))}{\|\mathbf{A}(P(x_0,I))\|}.  
\end{align}

\begin{align} \label{eq::10}
    \mathbf{A}(P(x_0,I)) = 
    \begin{bmatrix}
        \lvert I(x_0)-I(x_1) \lvert \\
         \lvert I(x_0)-I(x_2) \lvert \\
        \vdots \\
        \lvert I(x_0)-I(x_8) \lvert
    \end{bmatrix}.
\end{align}.

\noindent $I(x_i) \in [1,2,...,8]$ denotes locations relative to the central pixel $x_0$. $\mathbf{D}$ is a 8 component vector which is calculated using Eq. \ref{eq::9} and \ref{eq::10}. Our descriptor $\mathbf{D}$ is a simpler form of a descriptor proposed in \cite{trinh2019illumination} and it represents the normalized image gradient about the central pixel of patch $P(x_0)$, which was shown to be invariant to linear illumination changes: 

\begin{align} \label{eq::11}
{\bf D}(P(x + u,I)) = {\bf D}(aP(x,I) + b).
\end{align}

 

To preserve discontinuities at sharp object transitions and avoid staircasing artifacts in the in the calculated disparity map, we use a Huber function as a regularization term:

\begin{align} \label{eq::12}
E_S = \int_\Omega\lvert \nabla u \rvert_\epsilon dx,
\end{align}

\noindent where

\begin{align} \label{eq::13}
\left\lvert r \right\lvert_\epsilon =\begin{cases} 
\frac{r^2}{2\epsilon} &0\le |r| \le \epsilon, 
\\ |r|-\frac{\epsilon}{2} &\epsilon<|r|. \end{cases}.
\end{align}

\noindent and $\epsilon$ is a small positive constant. 


Finally, Eq. \ref{eq::6} takes the following form:

\begin{align} \label{eq::14}
    \min_u \left[\lambda \,\int_\Omega \left\lvert {\bf D}(P(x+u,I_t))-{\bf D}(P(x,I_s)) \right\lvert^2 dx + \int_\Omega\lvert \nabla u(x) \rvert_\epsilon dx \right],
\end{align}

\noindent which is solved by applying a primal-dual minimization scheme proposed in \cite{chambolle2011first}. As is common in most variational optical flow algorithms \cite{kroeger2016fast,werlberger2009anisotropic}, we use a coarse-to-fine warping framework to deal with large displacements. A scale factor of 0.5 is used to construct an image pyramid of $n$ levels. At each level, we perform $m$ warping iterations of optimizing energy functional Eq. \ref{eq::14}. In each level, the warping iteration is initialized with the current disparity field $u$, and a target image is warped towards the source image using the current disparity map.

\subsection{Experiment Setup}

\begin{table}
\scriptsize
\centering
\caption{Summary of parameters in our methods.}
\begin{tabular}{|c|c|c|}
\hline
 Parameter & Function & Value\\
\hline
$\alpha_s$ & Hyper-parameter in Eq. \ref{eq::1} & 20\\
$ \alpha_p$ & Hyper-parameter in Eq. \ref{eq::2} & 2\\
$th_{f}$ & Threshold to select outliers from the final confidence map & 0.5\\
$\lambda$ & Weight between data term and regularization term in Eq. \ref{eq::6} & 0.5 \\
$\epsilon$ & Hyper-parameter in Huber norm regularizer Eq. \ref{eq::7} & 0.1 \\
$m$ & Warping iterations at each image pyramid to solve Eq. \ref{eq::14} & 50 \\
$n$ & Levels of image pyramid to solve Eq. \ref{eq::14} & 4 \\
\hline
\end{tabular}
\label{tab:parameters}
\end{table}

We use several state-of-the-art learning-based methods, including PSMnet\cite{chang2018pyramid}, AAnet\cite{xu2020aanet}, LEAStereo\cite{cheng2020hierarchical}, and STTR\cite{li2021revisiting}, to generate raw disparity maps of images from the SERV-CT\cite{edwards2020serv} dataset and the SCARED dataset.\cite{allan2021stereo}. All the above learning methods are executed with their public models trained on the SceneFlow dataset\cite{flow}. We use the released light version of STTR to make it run on our local PC, of specifications listed further below.

We use all images in the SERV-CT dataset \cite{edwards2020serv} that includes 16 pairs of \textit{ex vivo} stereo endoscopic images of resolution 720 × 576. The dataset is collected from porcine full torso cadavers. Dense ground truth disparity maps and occlusion maps are computed from aligned CT scans.

We use five sub-datasets (dataset 1,2,3,7, and 8) of the SCARED dataset \cite{allan2021stereo}, collected \textit{ex vivo} from the abdominal anatomy of a porcine cadaver. We exclude the rest images in the sub-dataset, as they may contain intrinsic camera errors. Each sub-dataset contains five keyframes with associated camera calibration parameters, and the point cloud is reconstructed using structured light. In total, there are 25 pairs of stereo endoscopic images of resolution 1080 × 1024.

The proposed local disparity refinement method and global disparity refinement method are implemented in C++ and CUDA C++, which are running on an Intel i9-9900K CPU and a NIVIDA Titan X GPU, respectively. Parameters are shown in Table 1. Input images are resized to half the size of the original images. We examine the refinement performance of our proposed LDR and GDR and compare them to the closest work SDR \cite{yan2019segment} to ours. We used the released code of SDR with the default setting tuned to several datasets.

Root mean square disparity error (RMSE Disparity) and root mean square depth error (RMSE Depth) are used to evaluate errors between the estimated and ground truth results.

\section{Results}


\subsection{SERV-CT Dataset}

\subsubsection{Qualitative Evaluation}

\begin{figure}[htpb]%
\centering
\includegraphics[width=1.0\textwidth]{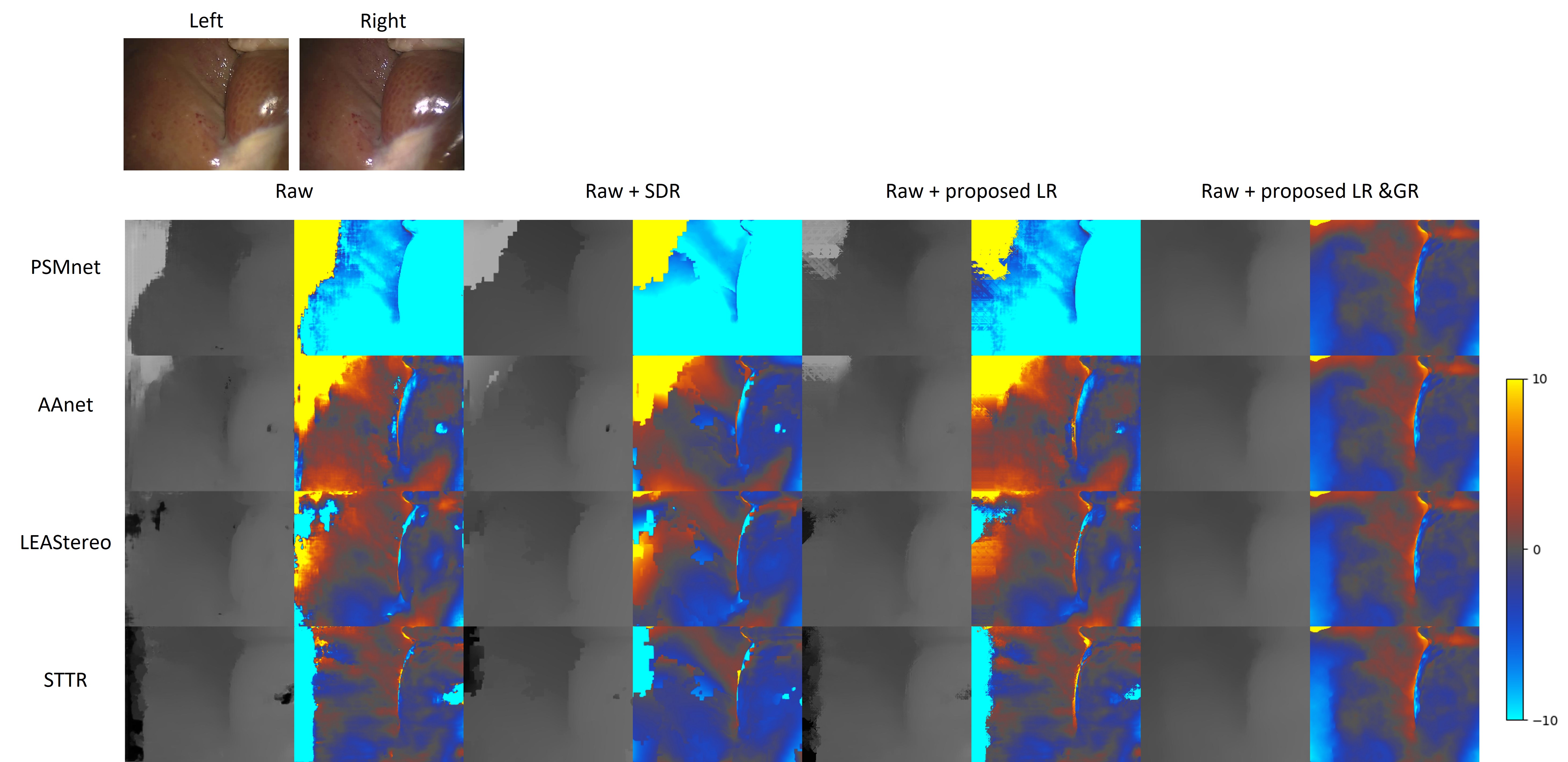}
\caption{Disparity maps of learning-based methods (PSMnet\cite{chang2018pyramid}, AAnet\cite{xu2020aanet}, LEAStereo\cite{cheng2020hierarchical}, STTR\cite{li2021revisiting}) and their refined results by SDR \cite{yan2019segment}, our LDR, and LDR + GDR, with disparity error maps compared with ground truth. Stereo endoscopic images and ground truth disparity map are shown on the top.}\label{fig1}
\end{figure}

\begin{figure}[htpb]%
\centering
\includegraphics[width=1.0\textwidth]{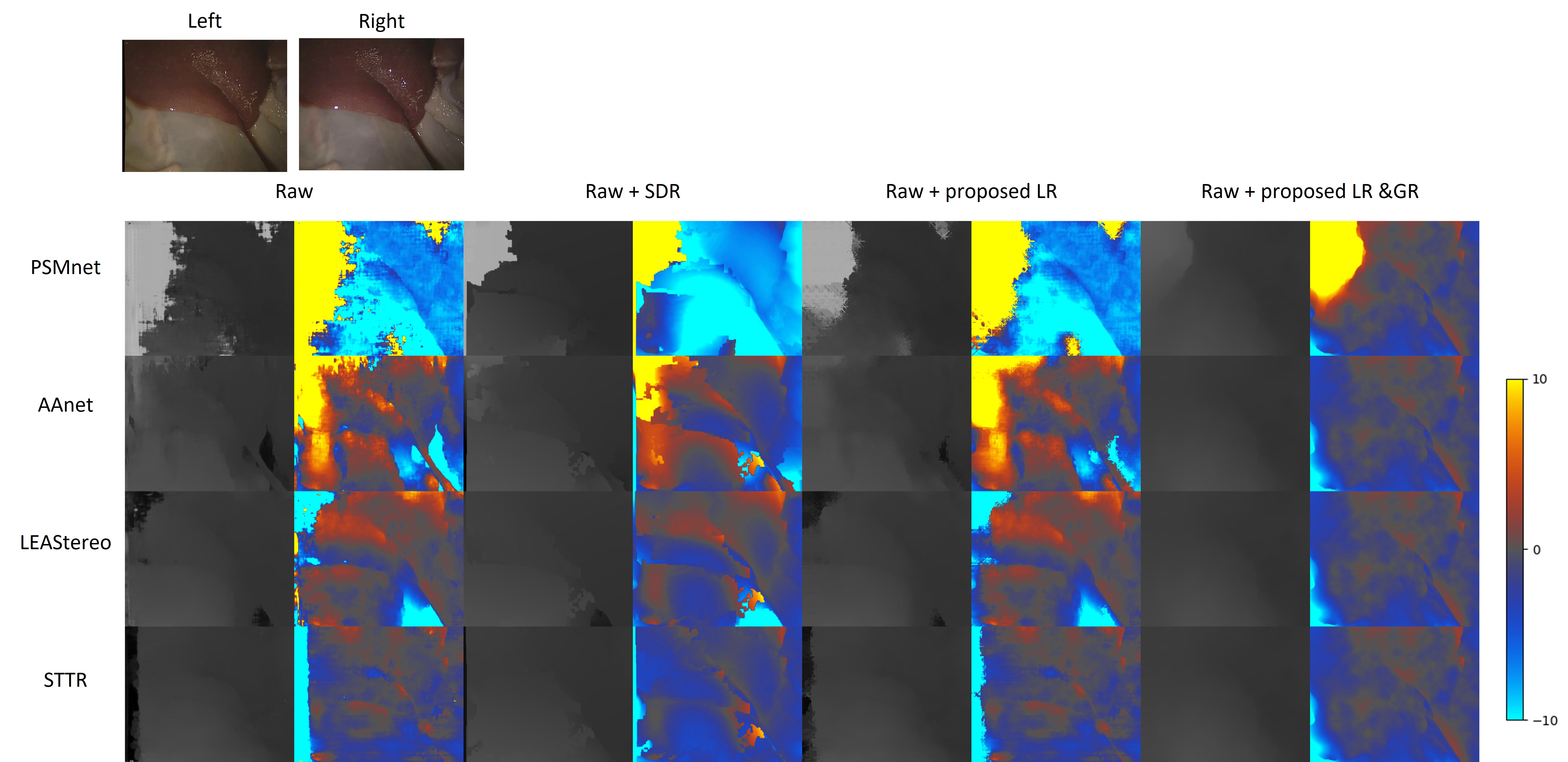}
\caption{Disparity maps of learning-based methods (PSMnet\cite{chang2018pyramid}, AAnet\cite{xu2020aanet}, LEAStereo\cite{cheng2020hierarchical}, STTR\cite{li2021revisiting}) and their refined results by SDR \cite{yan2019segment}, our LDR, and LDR + GDR, with disparity error maps compared with ground truth. Stereo endoscopic images and ground truth disparity map are shown on the top.}\label{fig2}
\end{figure}

Fig. \ref{fig1} and \ref{fig2} exhibit error images that indicate differences between the reference and predictions before and after refinement. 
We observe that raw disparity maps generated from the learning-based methods show significant disparity errors, especially in regions containing imaging artifacts such as specular highlights, occlusions, low texture and illumination differences. These error regions (i.e., the noise-corrupted regions) are characterized by spikes, strikes, and holes and are not continuous with their surrounding areas. These imaging artifacts are common in endoscopic images, but not in the natural images that they are trained on. Small error regions can be refined effectively via the proposed LDR, and SDR \cite{yan2019segment}. However, they are not able to refine large error-corrupted areas, such as errors around boundary occlusions. The proposed GDR could further improves the results refined by the LDR and, moreover, various imaging artifacts can also be effectively refined via the GDR.

\subsubsection{Quantitative Evaluation}

\begin{table*}[htpb]
    \centering
    \caption{Evaluation results on SERV-CT dataset. The statistical significance between the errors before refinement and after refinement is identified by $*(p<0.05)$.}
    \resizebox{0.99\linewidth}{!}{%
    \begin{tabular}{||c||c|c||c|c||c|c||}
    \hline
     & \multicolumn{2}{c||}{Occlusions included} 
     & \multicolumn{2}{c||}{Occlusions not included}  
    \\
    Method
      & RMSE Disparity (pixel)  & RMSE Depth (mm) 
     & RMSE Disparity (pixel)  & RMSE Depth (mm)   \\ \hline 
    PSMnet \cite{chang2018pyramid}  
    &   38.91  $\pm$   17.11  &   25.16  $\pm$    8.77  
    &   33.07  $\pm$   16.30  &   23.01  $\pm$    8.88\\
    PSMnet \cite{chang2018pyramid} + LDR  
    &   31.70  $\pm$   17.57  &   22.55  $\pm$    8.76  
    &   29.04  $\pm$   16.63  &   21.33  $\pm$    9.14    \\ 
    PSMnet \cite{chang2018pyramid} + LDR + GDR  
   &  $*$  8.17  $\pm$   10.38  &  $*$  7.89  $\pm$    7.85  
   &   $*$ 6.59  $\pm$    9.99  &   $*$ 6.48  $\pm$    7.69    \\
   \hline
    PSMnet \cite{chang2018pyramid} + SDR\cite{yan2019segment}  
   &   32.04  $\pm$   21.03  &   22.30  $\pm$    8.07  &   28.00  $\pm$   19.78  &   20.79  $\pm$    8.16     \\
   \hline
    AAnet \cite{xu2020aanet}  
    &   11.71  $\pm$    7.21  &   13.33  $\pm$    5.90  
    &    9.35  $\pm$    6.20  &   11.85  $\pm$    5.74  
    \\
     AAnet \cite{xu2020aanet} + LDR  
   &    9.39  $\pm$    7.13  &   10.04  $\pm$    6.72  
   &    7.53  $\pm$    6.29  &    8.39  $\pm$    6.32  \\
    AAnet \cite{xu2020aanet} + LDR + GDR  
   &  $*$  4.15  $\pm$    2.08  &  $*$  4.86  $\pm$    2.70  
   &  $*$  2.76  $\pm$    1.43  &  $*$  3.47  $\pm$    2.40  \\
    \hline
     AAnet \cite{xu2020aanet} + SDR\cite{yan2019segment}  
  &    9.22  $\pm$    4.92  &   11.36  $\pm$    4.87  
  &    6.98  $\pm$    3.80  &    9.51  $\pm$    4.60    \\
    \hline
    LEAStereo \cite{cheng2020hierarchical} 
   &    7.79  $\pm$    5.58  &   12.21  $\pm$    7.39  
   &    6.27  $\pm$    5.10  &   10.27  $\pm$    6.69      \\ 
     LEAStereo \cite{cheng2020hierarchical} + LDR
    &    6.06  $\pm$    4.66  &    9.66  $\pm$    6.54  
    &    4.49  $\pm$    3.89  &    7.05  $\pm$    5.57   
    \\  LEAStereo \cite{cheng2020hierarchical} + LDR + GDR
    &  $*$  3.96  $\pm$    1.79  &   $*$ 4.45  $\pm$    2.03  
    &  $*$  2.58  $\pm$    1.31  &   $*$ 3.06  $\pm$    1.73    \\ 
    \hline 
     LEAStereo \cite{cheng2020hierarchical} + SDR\cite{yan2019segment}
   &    5.05  $\pm$    2.83  &    7.24  $\pm$    3.68  
   &    4.17  $\pm$    2.38  &    6.42  $\pm$    3.56      \\ 
    \hline
    STTR \cite{li2021revisiting} 
   &   17.22   $\pm$    6.38  &   27.06   $\pm$    5.16  
   &    4.27   $\pm$    3.47  &    5.34   $\pm$    4.00    \\
    STTR \cite{li2021revisiting} + LDR
   &   13.23  $\pm$    6.25  &  $*$ 22.44  $\pm$    5.77 
   &    3.34  $\pm$    3.13  &    4.20  $\pm$    3.81
   \\ 
     STTR \cite{li2021revisiting} + LDR + GDR
   &  $*$  4.86  $\pm$    3.04  &   $*$ 5.97  $\pm$    3.57  
   &    2.95 $\pm$    1.68  &    3.36  $\pm$    1.70 
   \\ 
   \hline
    STTR \cite{li2021revisiting} + SDR\cite{yan2019segment}
   &   10.71  $\pm$    4.84  &   16.38  $\pm$    6.15  
   &    3.64  $\pm$    3.49  &    4.83  $\pm$    3.63  
   \\ 
   \hline
    \end{tabular}
    }%
    \label{tab:main}
\end{table*}

\begin{table*}[htpb]
    \centering
    \caption{Ablation Study of using different priors for GDR.}
    \resizebox{0.99\linewidth}{!}{%
    \begin{tabular}{||c||c||c||c||}
    \hline
     & \multicolumn{1}{c||}{Occlusions included} 
     & \multicolumn{1}{c||}{Occlusions not included}
    \\
     Method
      & RMSE Disparity (pixel)   
     & RMSE Disparity (pixel)     \\ \hline
    PSMnet\cite{chang2018pyramid}  + LDR + GDR  
   &     8.17  $\pm$   10.38  &     6.59  $\pm$    9.99  
   \\ 
    PSMnet\cite{chang2018pyramid}  + GDR   
    & 8.70 $\pm$ 9.98 & 6.94 $\pm$ 9.58  
    \\
   GDR (Level 4) 
    & 56.88 $\pm$ 24.85 & 56.39 $\pm$ 25.89 
   \\
   GDR (Level 6)
    & 8.38 $\pm$ 7.08 & 5.70 $\pm$ 4.75 
   \\
   \hline
    \end{tabular}
    }%
    \label{tab:ab}
\end{table*}

Table \ref{tab:main} further confirms our observations from the qualitative results and demonstrates the effectiveness and robustness of our proposed method. Decreases in errors are observed after each refinement stage, especially in the GDR stage. When including the occluded region in the evaluations, raw disparity maps estimated from LEAStereo\cite{cheng2020hierarchical} have the lowest 2D and 3D errors, with the RMSE $7.79 \pm 5.58$ pixel ($12.21 \pm 7.39$ mm). The errors are minimized to  $6.06 \pm 4.66$ pixel ($9.66 \pm 6.54$ mm) after LDR stage, and $ 3.96 \pm 1.79$ pixel ( $4.45 \pm 2.03$ mm) after GDR stage. 

All results from all networks have higher accuracy after excluding occluded regions, and STTR \cite{li2021revisiting} has the lowest error with $4.27 \pm 3.47$ pixel ($5.34 \pm 4.00 $ mm). The results of STTR\cite{li2021revisiting} can be further improved to $3.34 \pm 3.13$ pixel ($4.20 \pm 3.81$ mm) at the LDR stage, and $2.95 \pm 1.68$ pixel ($ 3.36 \pm 1.70$ mm) at the GDR stage.
Our method can also refine disparity maps predicted by PSMnet with significant errors. Excluding occluded regions, errors of PSMnet are refined from $ 31.70 \pm 17.57$ pixels to $6.59 \pm 9.99$ pixels.

We provide quantitative results in Table \ref{tab:ab} for the effects of using different priors for GDR and their running time. Different priors include using raw disparity maps from PSMnet, refined disparity maps after LDR, and no prior information.
We can observe that using better priors results leads to fewer errors in the final results. The errors associated with PSMnet as prior are higher than those associated with using refined disparity maps after LDR. LDR makes GDR more robust to error (noise)-corrupted priors. Without using priors from networks, the variational model of GDR requires an additional two-image pyramid level to obtain reasonable results. Priors from networks assist the variational model in achieving more accurate results. The refinement performance of LDR is comparable to SDR, and the refinement performance of GDR outperforms SDR (shown in Table \ref{tab:main}). 

The LDR and GDR ($n = 4$) are running at 0.118s and 0.375s per sample, which are more efficient than the SDR running at 1.794s per sample. GDR with $n = 6$ runs at 0.5s per sample.


\subsection{SCARED Dataset}

\subsubsection{Qualitative Evaluation}

\begin{figure}[htpb]%
\centering
\includegraphics[width=1.0\textwidth]{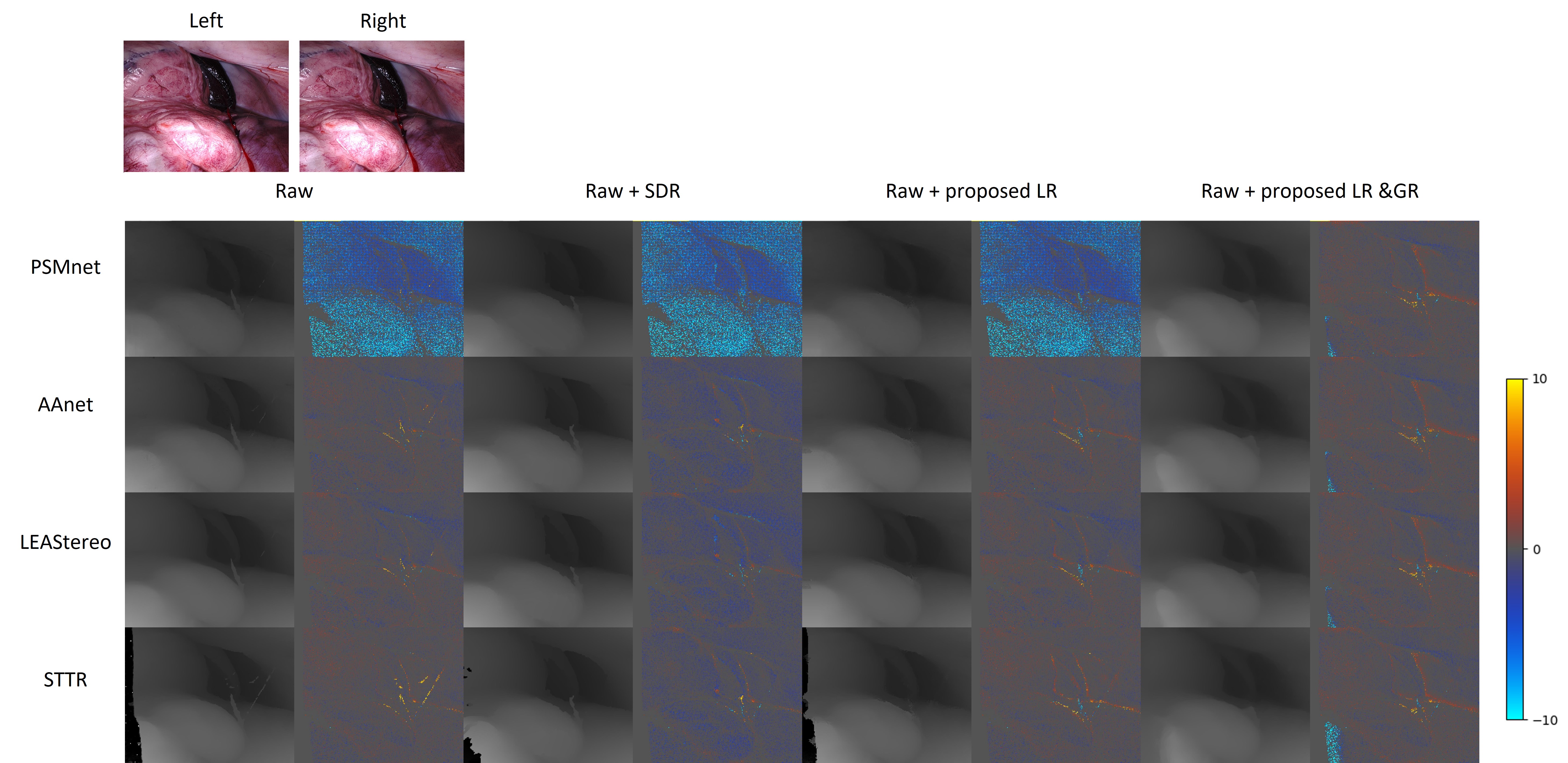}
\caption{Disparity maps of learning-based methods (PSMnet\cite{chang2018pyramid}, AAnet\cite{xu2020aanet}, LEAStereo\cite{cheng2020hierarchical}, STTR\cite{li2021revisiting}) and their refined results by SDR \cite{yan2019segment}, our LDR, and LDR + GDR, with disparity error maps compared with ground truth. Stereo endoscopic images and ground truth disparity map are shown on the top.}\label{fig3}
\end{figure}

\begin{figure}[htpb]%
\centering
\includegraphics[width=1.0\textwidth]{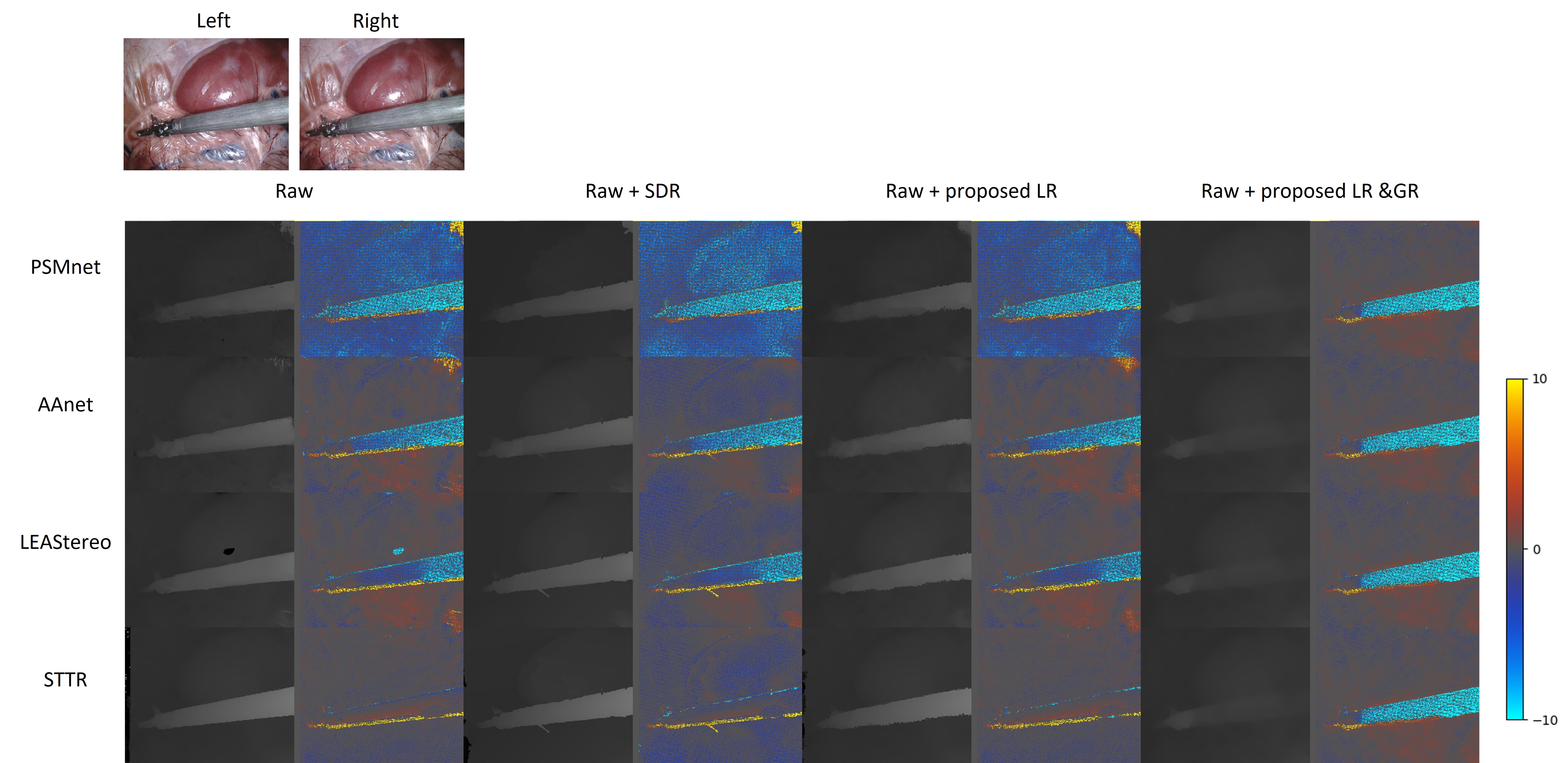}
\caption{Disparity maps of learning-based methods (PSMnet\cite{chang2018pyramid}, AAnet\cite{xu2020aanet}, LEAStereo\cite{cheng2020hierarchical}, STTR\cite{li2021revisiting}) and their refined results by SDR \cite{yan2019segment}, our LDR, and LDR + GDR, with disparity error maps compared with ground truth. Stereo endoscopic images and ground truth disparity map are shown on the top.}\label{fig4}
\end{figure}

Images in the SCARED dataset contain high texture with low noise, as shown in Fig. \ref{fig3} and Fig. \ref{fig4}. The SCARED dataset also includes surgical instruments not included in the SERV-CT dataset. In this cross-domain setting, the networks predict accurate disparity maps with clear boundaries that contain fewer artifacts than the SERV-CT dataset.
Disparity maps from PSMnet seem to lose scale, but can be refined correctly via the proposed GDR. From Fig. \ref{fig3}, proposed disparity refinement methods correct minor artifacts but do not deteriorate regions without artifacts. The proposed GDR has difficulty dealing with images that contain surgical instruments (Fig. \ref{fig4}). This may result from the weight of the smoothness term.

\subsubsection{Quantitative Evaluation}

\begin{table*}[htpb]
    \centering
    \caption{Evaluation results on SCARED dataset. The statistical significance between the errors before refinement and after refinement is identified by $*(p<0.05)$.}
    \resizebox{0.9\linewidth}{!}{%
    \begin{tabular}{||c||c||c||}
    \hline
    Method & RMSE Disparity (pixel)  & RMSE Depth (mm) \\
    \hline
      PSMnet \cite{chang2018pyramid}  
   &   13.71   $\pm$    6.33  &   13.77   $\pm$    3.48  \\
    PSMnet \cite{chang2018pyramid} + LDR  
   &   13.64  $\pm$    6.31  &   13.60  $\pm$    3.45    \\ 
    PSMnet \cite{chang2018pyramid} + LDR + GDR  
   &   $*$ 5.06  $\pm$    6.80  &   $*$ 3.40  $\pm$    2.67  \\
   \hline
    PSMnet \cite{chang2018pyramid} + SDR\cite{yan2019segment}  
   &   14.52  $\pm$    6.14  &   15.47  $\pm$    4.19     \\
   \hline
    AAnet \cite{xu2020aanet}  
   &    4.92  $\pm$    6.51  &    3.85  $\pm$    2.51   
    \\
     AAnet \cite{xu2020aanet} + LDR  
  &    4.63  $\pm$    6.44  &    3.35  $\pm$    2.45  \\
    AAnet \cite{xu2020aanet} + LDR + GDR  
  &    4.98  $\pm$    6.58  &    3.62  $\pm$    3.07  \\
    \hline
     AAnet \cite{xu2020aanet} + SDR\cite{yan2019segment}  
  &    5.13  $\pm$    6.29  &    4.11  $\pm$    2.52    \\
    \hline
    LEAStereo \cite{cheng2020hierarchical} 
   &    5.53  $\pm$    9.58  &    3.12  $\pm$    1.71      \\ 
     LEAStereo \cite{cheng2020hierarchical} + LDR
    &    5.47  $\pm$    9.59  &    2.85  $\pm$    1.44    
    \\  LEAStereo \cite{cheng2020hierarchical} + LDR + GDR
    &    5.68  $\pm$    8.80  &    3.24  $\pm$    2.63    \\ 
    \hline 
     LEAStereo \cite{cheng2020hierarchical}+SDR\cite{yan2019segment}
   &    6.07  $\pm$    9.54  &    3.76  $\pm$    1.74  \\ 
    \hline
    STTR \cite{li2021revisiting} 
   &    5.08  $\pm$    6.93  &    5.17  $\pm$    4.97  \\
    STTR \cite{li2021revisiting} + LDR
   &    4.76  $\pm$    6.92  &    4.53  $\pm$    4.31  
   \\ 
     STTR \cite{li2021revisiting} + LDR + GDR
   &    5.28  $\pm$    6.46  &    3.70  $\pm$    2.83
   \\ 
   \hline
    STTR \cite{li2021revisiting} + SDR\cite{yan2019segment}
   &    6.05  $\pm$    6.87  &    6.22  $\pm$    4.71 
   \\
   \hline
    \end{tabular}
    }%
    \label{tab:scared}
\end{table*}

Table \ref{tab:scared} suggests that, except for PSMnet, the networks can generalize well on the SCARED dataset. The RMSE disparity errors of AAnet, LEAStereo, and STTR are around 5 pixels. After LDR and GDR, disparity errors of PSMnet can be decreased from 13.71 $\pm$ 6.33 pixel to 5.06 $\pm$ 6.80 pixel, which are to the performance achieved on the other networks. LDR slightly improves the results, and GDR does not further improve or impair the results significantly. 

\section{Discussion}

We propose a disparity refinement framework consisting of local LDR and global GDR disparity refinement methods to maintain the robust performance of learning-based methods and apply them to medical images featured with limited data for training and various scenes. We conducted subjective and objective assessments of several state-of-the-art learning-based methods and refined results with our proposed LDR and GDR on the SERV-CT and SCARED datasets.

Images in the SERV-CT dataset feature a texture-less surface, specular highlights, and illumination differences, which pose difficulties for the generalization ability of learning-based methods. The proposed LDR and GDR can effectively correct artifacts, such as spikes, strikes, and holes. Leveraging assumptions to detect outliers and update them via interpolation, LDR opts for small noisy-corrupted regions. GDR using refined results as initialization from LDR can correct various types of errors. This initialization strategy improves the speed, accuracy, and robustness of the GDR model.

Compared with the SERV-CT dataset, the SCARED dataset has images with uniform (even) illumination and well-textured textures. The networks predict disparity maps with only a few errors. Although the statistical results are insignificant, the proposed LDR and GDR can refine the errors without damaging the pixels with accurate disparity values. 

The major limitation of GDR is its success in tackling images containing surgical instruments, due to smoothing of the instrument boundaries, which could be a side-effect of not having fully optimized the parameters utilized in these experiments. Nevertheless, to minimize boundary smoothing and improve the model performance on images featuring surgical instruments, as part of our future work, we will investigate the use of variational models \cite{zach2007duality,werlberger2009anisotropic} that feature image gradient-based anisotropic weighting of the regularized.

\section{Conclusion}\label{sec13}

We have presented a robust and accurate disparity refinement framework with a local disparity method and a global disparity refinement method for learning-based methods in a cross-domain setting. In cases where networks perform poorly, our disparity refinement methods can improve the accuracy of the disparity maps. Moreover, our disparity refinement methods can improve the disparity maps of networks when they are corrupted by noise and without compromising network performance when networks generalize well on new domains. 

\section{Acknowledgement}
Research reported in this publication was supported by the National Institute of General Medical Sciences of the National Institutes of Health under Award No. R35GM128877 and by the Office of Advanced Cyber-infrastructure of the National Science Foundation under Award No.1808530.
\section{Disclosure}
Nothing to disclose.



\bibliography{sn-bibliography}

\end{document}